\def\year{2019}\relax
\documentclass[letterpaper]{article} %DO NOT CHANGE THIS
\usepackage{aaai19}  %Required
\usepackage{times}  %Required
\usepackage{helvet}  %Required
\usepackage{courier}  %Required
\usepackage{url}  %Required
\usepackage{graphicx}  %Required
\usepackage{latexsym}
\usepackage{amsmath,amsfonts,amssymb}
\usepackage{symbols}
\usepackage{multirow}
\usepackage{booktabs}
\usepackage{ulem}

\newcommand{\newcite}[1]
{\citeauthor{#1}~\shortcite{#1}}

\begin{document}
\title{Knowledge Based Machine Reading Comprehension}
\author{Yibo Sun$^\S$\thanks{\ \ Work is done during internship at Microsoft Research Asia.},\ \
     Daya Guo$^{\natural*}$,
     Duyu Tang$^\ddag$,
     Nan Duan$^\ddag$,
     Zhao Yan$^{\dag*}$,
	\bf Xiaocheng Feng$^\S$, Bing Qin$^\S$\\
	$^\S$Harbin Institute of Technology, Harbin, China\\
	$^\ddag$Microsoft Research Asia, Beijing, China \\
	$^\natural$Guangdong Key Laboratory of Big Data Analysis and Processing, Guangzhou, China\\
    $^\dag$Beihang University, Beijing, China\\
	{\small \tt \{ybsun,xcfeng,qinb\}@ir.hit.edu.cn}\\
	{\small \tt \{dutang,nanduan\}@microsoft.com}\\
    {\small \tt \{guody@5mail2\}@sysu.edu.cn}
    {\small \tt \{yanzhao\}@buaa.edu.cn}\\
}
\maketitle
\begin{abstract}
Machine reading comprehension (MRC) requires reasoning about both the knowledge involved in a document and knowledge about the world.
However, existing datasets are typically dominated by questions that can be well solved by context matching, which fail to test this capability.
To encourage the progress on knowledge-based reasoning in MRC, we present knowledge-based MRC in this paper, and build a new dataset consisting of 40,047 question-answer pairs.
The annotation of this dataset is designed so that successfully answering the questions requires understanding and the knowledge involved in a document.
We implement a framework consisting of both a question answering model and a question generation model, both of which take the knowledge extracted from the document as well as relevant facts from an external knowledge base such as Freebase/ProBase/Reverb/NELL.
Results show that incorporating side information from external KB improves the accuracy of the baseline question answer system.
We compare it with a standard MRC model BiDAF, and also provide the difficulty of the dataset and lay out remaining challenges.
\end{abstract}

\section{Introduction}

In recent years, there has been an increasing interest in Machine reading comprehension (MRC),
which plays a vital role in the assessment of how well a machine could understand natural language.
Several datasets~\cite{rajpurkar2016squad,onishi2016did,hill2015goldilocks} for machine reading comprehension have been released in recent years and have driven the evolution of powerful neural models.
However, much of the research up to now has been dominated by answering questions that can be well solved solved using superficial information, yet struggles to do accurate natural language understanding and reasoning.
For example, \newcite{jia2017Adversarial} show that existing machine learning systems for MRC perform poorly under adversarial evaluation.
Recent developments in MRC datasets~\cite{Kocisky2018NarrativeQA,Rajpurkar2018SQUAD2,Welbl2018MultiHopRC} have heightened the need for deep understanding.

Knowledge has a pivotal role in accurately understanding and reasoning natural language in MRC. Previous research~\cite{hirsch2003reading,carrell1983three} has established that human reading comprehension requires both words and world knowledge. In this paper, we consider words and world knowledge in the format of triplets (subject, predicate, object).
Specifically, we believe the advantages of using knowledge in MRC are three-fold.
\textbf{First}, utilizing knowledge in MRC supports reasoning over multiple triplets because a single triplet may not cover the entire question.
Multi-hop reasoning is also a long-standing goal in question answering.
\textbf{Second}, building a question answering system based on triplet-style knowledge facilitates the interpretability of the decision making process.
Triplets organize the document together with KBs as a graph, where a well-designed model such as PCNet, which we will describe in a later section, expressly reveal rationales for their predictions.
\textbf{Third}, representing the documents as knowledge allows for ease of accessing and leveraging the knowledge from external/background knowledge because the knowledge representation of a document is easily consistent with both manually curated and automatically extracted KBs.

In this paper, we present knowledge based machine reading comprehension, which requires reasoning over triplet-style knowledge involved in a document. However, we find published dataset do not sufficiently support this task. We conduct preliminary exploration on SQuAD~\cite{rajpurkar2016squad}. We use a strong open IE algorithm~\cite{clausie} to extract triplets from the documents and observe that only 15\% of instances have an answer that is exactly the same as the corresponding subject/object in the extracted triplets.
To do knowledge-based MRC, We build a new dataset consisting of 40,047 examples for the knowledge based MRC task. The annotation of this dataset is designed so that successfully answering the questions requires understanding and the knowledge involved in a document.
Each instance is composed of a question, a set of triplets derived from a document, and the answer.

We implement a framework consisting of both a question answering model and a question generation model, both of which take the knowledge extracted from the document as well as relevant facts from an external knowledge base such as Freebase/ProBase/Reverb/NELL. The question answering model gives each candidate answer a score by measuring the semantic relevance between representation and the candidate answer representation in vector space. The question generation model provides each candidate answer with a score by measuring semantic relevance between the question and the generated question based on the semantics of the candidate answer.
We implement an MRC model BiDAF~\cite{seo2016bidirectional} as a baseline for the proposed dataset.
To test the \mbox{scalability} of our approach in leveraging external KBs,
we use both manually created and automatically extracted KBs, including
Freebase~\cite{bollacker2008freebase}, ProBase~\cite{wu2012probase}, NELL~\cite{carlson2010toward} and
Reverb~\cite{fader2011identifying}.
Experiments show that incorporating evidence from external KBs improves both the matching-based
and question generation-based approaches.
Qualitative analysis shows the advantages and limitations of our approaches, as well as the remaining challenges.

\section{Task Definition and Dataset}
We formulate the task of knowledge based machine reading comprehension, which is abbreviated as KBMRC, and
describe the dataset built for KBMRC and the external open KBs leveraged in this work.

\begin{figure}[h]
	\centering
	\includegraphics[width=.47\textwidth]{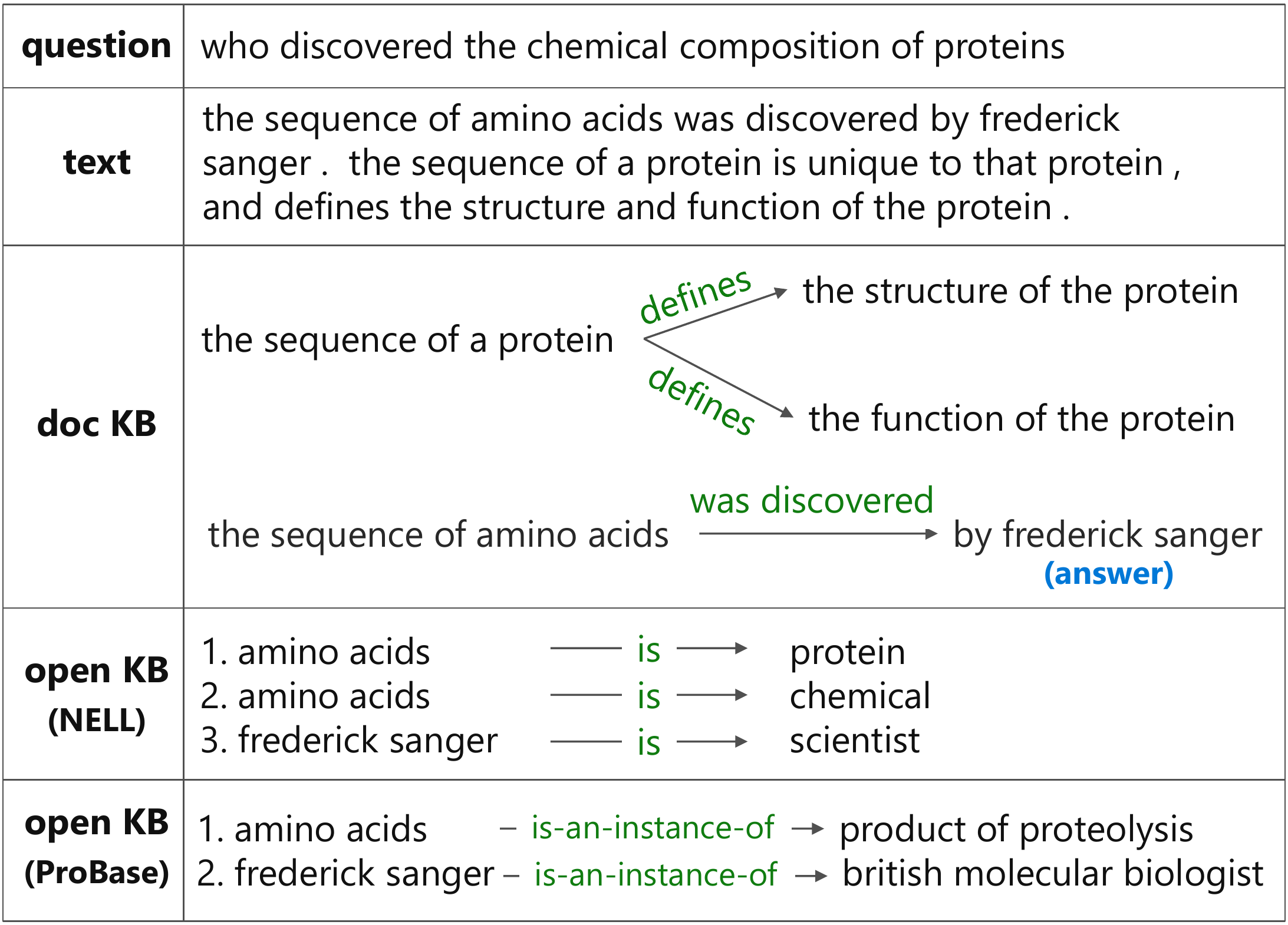}
	\caption{An example that illustrates the task.}
	\label{fig:task-definition}
\end{figure}

\paragraph{Task Definition}
The input of KBMRC includes a natural language question $q$,
a knowledge base $kb^d$ derived from the document $d$, and potentially an external knowledge base $kb^e$.
Both $kb^d$ and $kb^e$ consist of a set of triplets $\{f_1,..f_i, ..f_n\}$,
in which a triplet $f_i$ is composed of a subject $sbj_i$, a predicate $pred_i$,
and one or more arguments $arg1_i$, $arg2_i$, etc.
The output is an subject or an argument from $kb^d$ which correctly answers the question $q$.
Figure~\ref{fig:task-definition} gives an example to illustrate the task.

\paragraph{Dataset Construction}
We build our dataset upon WebAssertions~\cite{yan2018assertion}, which is valuable to us as each sample of in this dataset contains a document-question pair and triplets extracted from the document by an open IE algorithm~\cite{clausie}.
Questions are collected from the search log of a search engine and documents are collected from a commercial search engine's search result.
However, they do not annotate which subject/argument from a triplet is the answer. Based on this consideration, we make
further annotations in order to point out which subject/argument is the correct answer. To reduce annotation efforts, we take every argument from a correct triplet as the candidate answer and provide each annotator with a question and a document. Annotators are asked to determine whether a subject/argument is a correct answer or not. We assign each candidate
to three annotators and only collect the instances that are labeled as correct by no less than two annotators.

Statistics of the dataset are given in Table~\ref{tab:stat-dataset}.
The distribution of questions in our dataset is given in Table~\ref{tab:stat-question-type}.

\begin{table}[h]
	\centering
	\small
	%	\begin{tabular}{cc}
	\begin{tabular}{l|c}
		\hline
%		\textbf{KB} & \textbf{\#Triples} \\
%		\hline
		\# of training instances & 30,461 \\
		\# of dev instances & 4,481 \\
		\# of test instances & 5,105 \\
		\hline
		Avg. triplets / document & 5.85 \\
		Avg. arguments / document &  14.27\\
		Avg. predicates / document & 5.85 \\
		Avg. words / argument & 3.14 \\
		Avg. words / predicate & 1.31 \\
		Avg. words / question & 5.95 \\
		\hline 		
	\end{tabular}
	\caption{Statistics of the dataset we create for KBMRC.}\label{tab:stat-dataset}
\end{table}

To measure the complexity of the dataset, we use a rule-based approach to detect the anchor
and measure the coverage of correct answers through 1-hop and 2-hop paths.
The details of how we get the anchor and the paths are described in Section Approach Overview.
We observe that the coverage of 1-hop candidates is 55.6\% and that of 2-hop candidates is 69.6\%.
This indicates that deep/multi-hop inference is required in this dataset.
\begin{table}[h]
	\centering
    \small
%	\begin{tabular}{cc}
		\begin{tabular}{c|c|c|c|c|c|c}
			\hline
			\textbf{Type} & who& what& where& when& how& other\\
			\hline
			\textbf{\#} & 5815& 6,377 & 5,860& 3,423& 8,628& 9,944\\
			\textbf{\%}  & 14.5\%& 15.9\% & 14.6\%& 8.5\%& 21.5\%& 24.8\%\\
		\hline 		
		\end{tabular}
	\caption{Distribution of questions in our dataset.}\label{tab:stat-question-type}
\end{table}

We believe that one advantage of KBMRC is that the form of knowledge representation
of a document is analogous to that of large-scale knowledge bases.
To explore the scalability of our approach, we investigate our publicly available \mbox{KBs}
which have been used for open question answering.
Statistical information on the open KBs are given in ~\newcite{fader2014open},
from which we can find that these KBs include both manually curated ones (e.g. Freebase) and
the ones that are automatically extracted from web documents (e.g. Reverb, ProBase, NELL).
We use each open KB individually in the experiment and report the results in the next section.

\section{Approach Overview}
Our framework consists of a question answering model and a question generation model.

We implement two question answering models, which directly measure the semantic similarity between questions and candidate answers in the semantic space.
First, to make the model's prediction more explainable, we implement a path based QA model PCNet.
In this model, to get candidate answers from the triplets based document for a given question.
We first retrieve the an ``anchor'' point\footnote{The anchor point is a subject/object mentioned in the question.} $arg1_i$ or $arg2_i$ in the document fact $f_i$.
These anchors are selected based on edit distance between the words in the questions and the arguments.
Then we regard all arguments in the 1-hops and 2-hops fact of the anchors as answers.
However, the coverage of the candidate answers can be 100\% in the model. We then implement a second end-to-end neural model KVMenNet which covers all the answers but with less interpretability. Both models generate a score $f_{qa}(q, a)$ of each candidate answer.

We then implement a generation-based model. The motivation to design this model is that we want to associate natural language phrases with knowledge based representation.
It takes semantics of a candidate answer as the input and generates a question $\hat{q}$.
Then a paraphrasing model gives a score $f_{qg}(q,\hat{q})$, which is computed between
the generated question $\hat{q}$ and the original question $q$
, as the ranking score.

We get the final scores $S(q,a)$ used for ranking as follows.
\begin{align}
	S(q,a) = \lambda f_{qa}(q, a) + (1-\lambda) f_{qg}(q,\hat{q})
\end{align}

\begin{figure}[h]
	\centering
	\includegraphics[width=.40\textwidth]{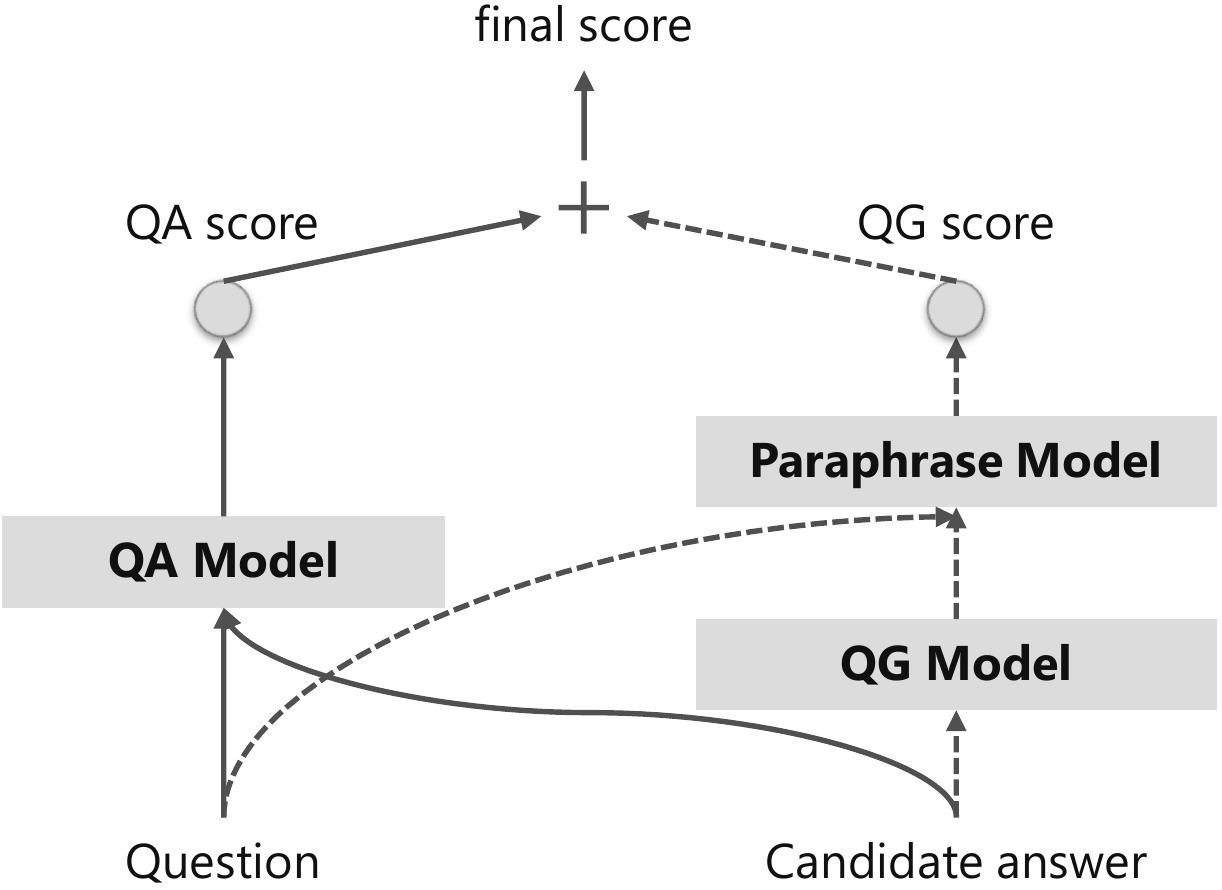}
	\caption{Approach overview.}
	\label{fig:approach-overview}
\end{figure}

Moreover, we incorporate side information from external KBs into the three models.
The details of how we use external KBs to enhance the representation of elements in the document KB will
be described in section Incorporating External Knowledge.

\section{The Question Answering Model}
\label{section:matching_models}
In this section, we present two matching \mbox{models} to measure the
semantic relatedness between the question and the candidate answer in the vector semantic space.
Afterwards, we introduce our strategy that incorporates open KB as external knowledge to enhance
both models.

\subsection{QA Model 1: PCNet}
\label{section:pc_nets}
We follow \newcite{bordes2014question,dong-EtAl:2015:ACL-IJCNLP1} and develop PCNet, which is short for path- and context- based neural network.
In PCNet, candidate answers come from  arguments of the document KB, and each candidate answer is represented with its neighboring arguments and predicates as well as its path from the anchor in the document \mbox{KB}.
We use a rule-based approach based on string fuzzy match to detect the anchor.
Each argument is measured by $\sum_{i,j}^{}\mathbb{I}(arg_i, q_j)$, where $i$ and $j$ iterate across argument words and question words, respectively.
$\mathbb{I}(x,y)$ is an indicator function whose value is 1 if the minimum edit distance between $x$ and $y$ is no more than 1, otherwise it is 0.
The arguments linked to the anchor with 1-hop and 2-hop paths are regarded as candidate answers.
Since an argument might include multiple words, such as ``\textit{the popular angry bird game}'', we use GRU based RNN to
get the vector representation of each argument/predicate.
The path representation $v_{p}$ is computed by averaging the vectors of elements in the path.
Similarly, we use another RNN to get the vector of each neighboring argument/predicate, and average them to get the context vector $v_{c}$.

We represent the question $q$ using a bidirectional RNN layer based on the GRU unit. The concatenation of the last hidden vectors from both directions is used as the question vector $v_{q}$.
The dot product is used to measure the semantic relevance between the b  b question and two types of evidence.
\begin{align}
	f_{qa}(q, a) = v_{q}^{T}v_{p} + v_q^{T}v_{c}
\end{align}

\subsection{QA Model 2: KVMemNet}
Despite the interpretability of PCNet, the coverage of anchor detection limits the upper bound of the approach.
We implement a more powerful method based on the key-value memory network \cite{miller2016key}, KVMemNet for short, which has proven powerful in KB-based question answering.

The KVMenNet could be viewed as a ``soft'' matching approach, which includes a parameterized memory consisting of key-value pairs.
Intuitively, keys are used for matching with the question, and values are used for matching to the candidate answer.
Given a KB fact ($subj, pred, obj$), We consider both directions and add two key-value pairs in the memory, namely ($key = subj +pred, value=obj$) and ($key =obj + pred, value=subj$).
The vectors of arguments/predicates are calculated the same way as described in PCNet.
The concatenation of two vectors is used as the key vector $v_{key}$.

Each memory item is assigned a relevance probability by comparing the question to each key.
\begin{equation}
\alpha_{key_i} =  {softmax}(v_q \cdot v_{key_i})
\end{equation}
Afterwards, vectors in memory ($v_{value}$) are weighted summed according to their addressing probabilities, and the vector $v_o$ is returned.
\begin{equation}
v_o = \sum_i \alpha_{key_i}v_{value_i}
\end{equation}
In PCNet, reasoning over two facts is achieved by incorporating 2-hop paths, while in KVMemNet this is achieved by repeating the memory access process twice.
After receiving the result $v_o$, we update the query with $q_2 = R(v_q + v_o)$, where $R$ is model parameter.
Finally, after a fixed number $n$ hops ($n=2$ in this work), the resulting vector is used to measure the relevance to candidate answers via the dot product.

\subsection{Training and Inference}
Let $\mathcal{D} = \{(q_i, a_i); i = 1,
\ldots, |\mathcal{D}|\}$ be the training data consisting of
questions $q_i$ paired with their correct answer $a_i$.
We train both matching models with a margin-based ranking loss function, which is calculated as follows, where $m$ is the margin (fixed to $0.1$) and $\bar{a}$ is randomly sampled from a set of incorrect candidates $\bar{\mathcal{A}}$.
\begin{equation} \label{eq:loss}
\sum_{i=1}^{|\mathcal{D}|} \sum_{\tiny{\bar{a} \in \bar{\mathcal{A}}(q_i)}}
\max\{0, m - S(q_i,a_i) + S(q_i,\bar{a})\},
\end{equation}

 For testing,  given a question $q$, the model predicts the answer based on the following equation, where $ \mathcal{A}(q)$ is the candidate answer set.
\begin{equation}\label{eq:inf}
\hat{a} = {\mbox{argmax}}_{a' \in \mathcal{A}(q)} S(q,a')
\end{equation}

\section{The Question Generation Model}
\label{section:generation_models}
In this section, we present the generation model which generates a question based on the semantics
of a candidate answer. Afterward, we introduce how our paraphrasing model, which measures the semantic
relevance between the generated question and the original question, is pretrained.

\subsection{Hierarchical seq2seq Generation Model}
Our question generation model takes the path between an ``anchor'' and the candidate answer, and outputs a question.
We adopt the sequence to sequence architecture as our basic question generation model due to its effectiveness in natural language generation tasks.
The hierarchical encoder and the hierarchical attention mechanism are introduced to take into account the structure of the input facts.
Facts from external KBs are conventionally integrated into the model using the same way as described in the matching model.

\begin{figure}[h!]
	\centering
	\includegraphics[width=.48\textwidth]{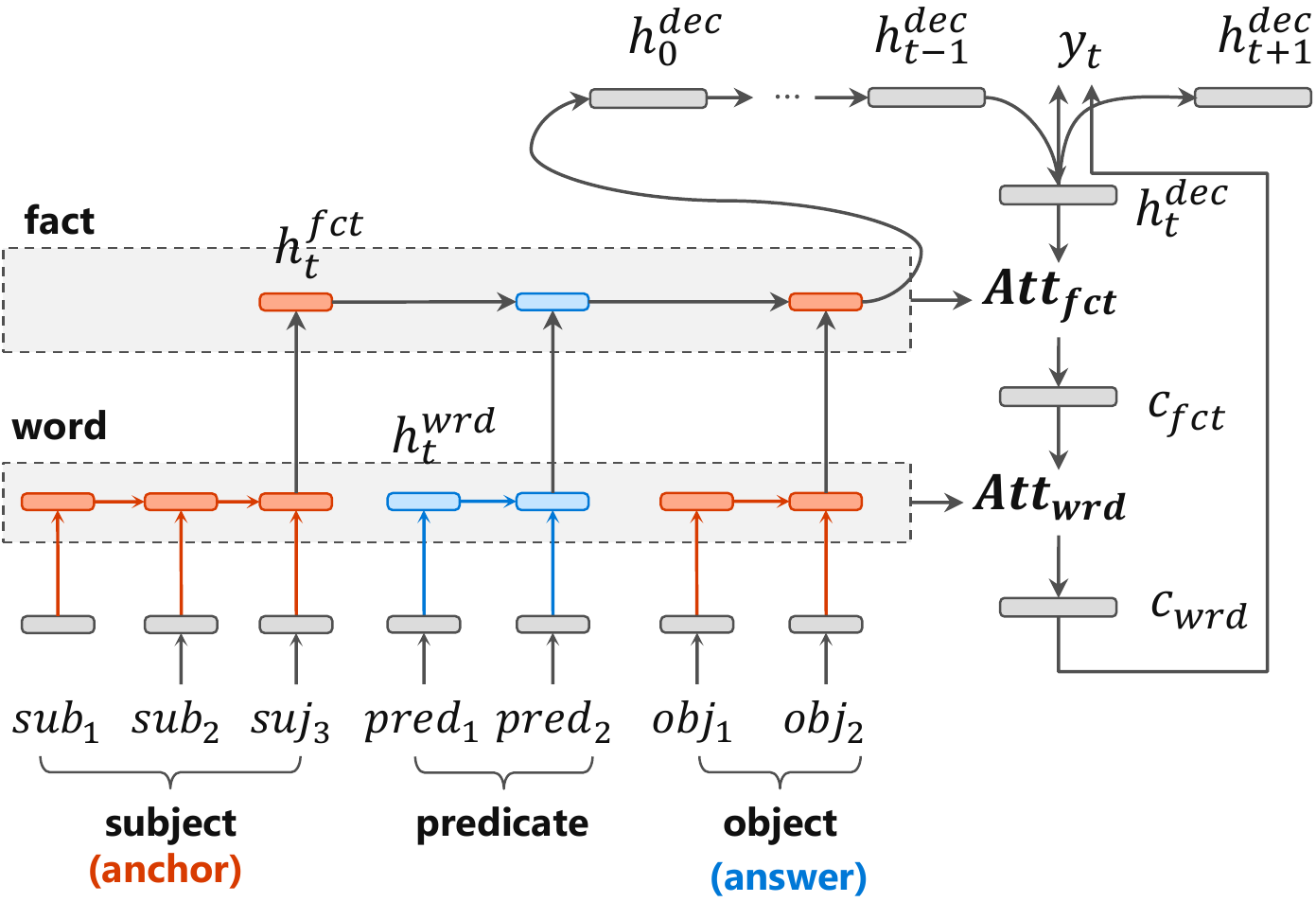}
	\caption{The question generation model.}
	\label{fig:qg}
\end{figure}
As illustrated in Figure~\ref{fig:qg}, the question generation model contains an encoder and a decoder.
We use a hierarchical encoder consisting of two \mbox{layers} to model the meaning of each element (subject, predicate or object) and the relationship between them.
Since each element might contain multiple words, we use RNN as the word layer encoder to get the representation of each element.
We define the fact level sequence as a path starting from the anchor and ending at the candidate answer.
Another RNN is used as the fact level encoder.
The last hidden state at the fact layer is fed to the decoder.

We develop a hierarchical attention mechanism in the decoder, which first makes soft alignment over the hidden states at the fact layer, the output of which is further used to attend to hidden states at the word level.
Specifically,
given the decoder hidden state $h^{dec}_{t-1}$, we use a fact-level attention model $Att_{fct}$ to calculate the contextual vector, which is further combined with the current hidden state $h^{dec}_{t}$, resulting in  $c_{fct}$.
The contextual vector is calculated through weighted averaging over hidden vectors at the fact level, which is given as follows, where $\odot$ is dot product, $h^{fld}_j$ is the $j$-th hidden state at the fact layer from the encoder, and $l_f$ is the number of hidden states at the fact level.
\begin{align}
c_{fct} &= GRU(h^{dec}_{t}, \sum_{j=1}^{{l_f}} \alpha_{tj}h^{fct}_j) \\
\alpha_{tj} &= \frac{ exp(h^{dec}_{t-1} \odot h^{fct}_j)}  {\sum_{k=1}^{{l_f}}exp(h^{dec}_{t-1} \odot h^{fct}_k)}
%e_{tj} &= a(h^{dec}_{t-1}, h^{fld}_j)
\end{align}
Similarly, we feed $c_{fct}$ to the word-level attention function $Att_{wrd}$ and calculate over hidden vectors at the word-level.
The output $c_{wrd}$ will be concatenated with $h^{dec}_{t}$ to predict the next word.

Since many entity names of great importance are rare words from the input, we use the copying mechanism \cite{gu-EtAl:2016:P16-1}
that learns when to replicate words from the input or to predict words from the target vocabulary.
The probability distribution of generating the word $y$ is calculated as follows, in which the $softmax$ function is calculated over a combined logits from both sides.
\begin{equation}\label{equa:copy distribution}
\begin{split}
&p(y)=\frac{exp({e_y \odot W_g[h_t^{dec};c_{wrd}]})+exp(s_c(y))}{Z}\\
&s_c(y) = c_{wrd} \odot{tanh(W_ch_t^{wrd})}
\end{split}
\end{equation}
We train our question generation model with maximum likelihood estimation. The loss function is given as follows, where $D$ is the training corpus.
We use beam search in the inference process.
\begin{equation}
l = -\sum_{(x,y)\in D} \sum_t log p(y_t |y_{<t}, x)
\end{equation}
An advantage of the model is that external knowledge could be easily incorporated with the same mechanism as we have described
in section Incorporating External Knowledge.
We enhance the representation of an argument or a predicate by concatenating open KB vectors into encoder hidden states.

\subsection{The Paraphrasing Model}
The paraphrasing model is used to measure the semantic relevance between the original question and the question generated from the QG model.
We use bidirectional RNN with gated recurrent unit to represent two questions, and compose them with element-wise multiplication.
The results are followed by a $softmax$ layer, whose output length is 2. The model is trained by minimizing the cross-entropy error, in which the supervision is provided in the training data.

We collect two datasets to train the paraphrasing model.
The first dataset is from Quora dataset\footnote{\url{https://data.quora.com/First-Quora-Dataset-Release-Question-Pairs}}, which is built for detecting whether or not a pair of questions are semantically equivalent.
345,989 positive question pairs and 255,027 negative pairs are included in the first dataset.
The second dataset includes web queries from query logs, which are obtained by clustering the web queries that click the same web page.
In this way we obtain 6,118,023 positive query pairs. We implement a heuristic rule to get 6,118,023 negative instances for the query dataset.
For each pair of query \mbox{text}, we clamp the first query and retrieve a query that is mostly similar to the second query.
To improve the efficiency of this process, we randomly sample 10,000 queries and define the ``similarity'' as the number of  co-occurred words between two questions.
During training, we initialize the values of word embeddings with $300d$ Glove vectors\footnote{\url{https://nlp.stanford.edu/projects/glove/}}, which is learned on Wikipedia texts.
We use a held-out data consisting of 20K query pairs to check the performance of the paraphrasing model.
The accuracy of the paraphrasing model on the held-out dataset is 87.36\%.
\section{Incorporating External Knowledge}
\label{section:add_knowledge}
There are many possible ways to implement the idea of improving question answering with external KB.
In this work, we use external \mbox{KBs} (such as NELL and ProBase) to enhance the representations of elements in the document KB.
For instance, the argument  ``\textit{the sequence of amino \mbox{acids}}'' in Figure \ref{fig:task-definition} from the document KB retrieves (``\textit{amino acids}'', `\textit{is}'', ``\textit{protein}'') from \mbox{NELL}.
Enhanced with this additional clue, the original argument is a better match to the question.

Similar to \newcite{khot2017answering}, we use ElasticSearch\footnote{https://www.elastic.co/products/elasticsearch} to retrieve facts from open KBs.
We remove stop words from tokens of each argument and predicate in a document KB and regard the remained words as ElasticSearch queries.
We set different search option for arguments and predicates, namely setting arguments as the \mbox{dominant} searchable fields for argument queries and setting predicates as the \mbox{dominant} searchable fields for predicate queries.
We save the top 10 hints and selectively use them in the experiment.

We regard the retrieved facts from the external \mbox{KB} as neighbors to the arguments to be enhanced.
Inspired by \newcite{scarselli2009graph}, we update the vector of an element $v_e$ as follows, where
$\inc(e)$ and $\out(e)$ represent adjacent arguments from the facts retrieved by object and subject, respectively.
In this work, $f(\cdot)$ is implemented by averaging the vectors of two arguments.
\begin{equation}
v_e=v_e + \sum_{o' \in \inc(e)}^{}f(v_{p'},v_{s'})+\sum_{s' \in \out(e)}^{}f(v_{p'},v_{o'})
\end{equation}

\begin{figure*}[t]
	\centering
	\includegraphics[width=.95\textwidth]{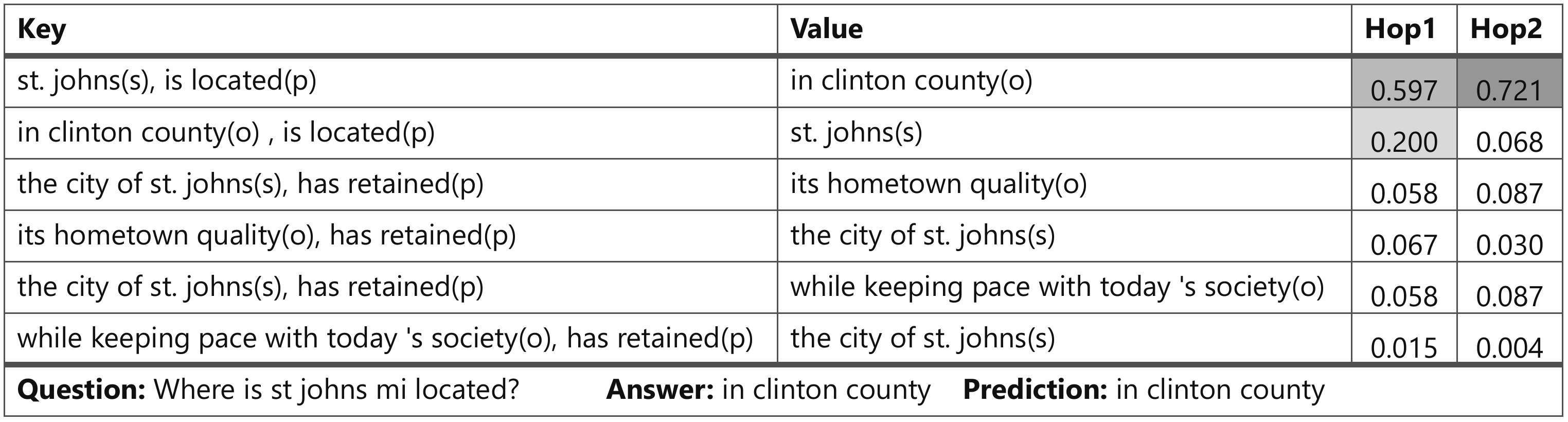}
	\caption{Model output of KVMemNet on the dev set. We show key-value pairs in the memory, and the probabilities of each hop used during inference. (s), (p) and (o) stand for subject, predicate, and object, respectively.}
	\label{fig:kvmn_sample_2hops}
\end{figure*}

\section{Related Work}
The task of KBMRC differs from machine reading comprehension (MRC) in both input and output aspects.
The input of KBMRC is the knowledge including both word knowledge extracted from the document and world knowledge retrieved from external knowledge base,
while the input of MRC is the unstructured text of a document.
The output of KBMRC is a subject or an argument, while the output in MRC is a text span of the document.
Meanwhile, KBMRC facilitates the accessing and leveraging of knowledge from external KBs because the document KB is consistent with the representation of facts in external KBs.

KBMRC also relates to knowledge-base question answering (KBQA) \cite{yih2015semantic}, which \mbox{aims} to answer questions based on an external large-scale KB such as Freebase or ProBase.
KBMRC differs from KBQA in that the original KB comes from the content of a document.
External KB is used in this work to enhance the document KB.
Moreover, existing benchmark datasets for KBQA such as WebQuestions \cite{berant2013semantic} are typically limited to simple questions.
The KBMRC task requires reasoning over two facts from the document KB.

Our approach draws inspiration from two main classes in existing approaches of KBQA,
namely ranking based and parsing based.
Ranking based approaches \cite{bordes2014question,berant2014semantic} are bottom-up,
which typically first find a set of candidate answers and then rank between the candidates
with features at different levels to get the answer.
Parsing-based approaches \cite{fader2014open} are top-down,
which first interpret logical form from a natural language utterance, and then do execution to yield the answer.
Ranking-based approaches achieve better performances than parsing-based approaches on WebQuestions,
a benchmark dataset for KBQA. We follow ranking-based approaches,
and develop both a matching-based model with features at different levels and a question generation model.
More references can be found at {\url{https://aclweb.org/aclwiki/Question_Answering_(State_of_the_art)}}.

Our work also relates to~\newcite{khot2017answering},
which uses open IE outputs from external text corpora to improve multi-choice question answering.
However, our work differs from them in that their task does not contain document information.
Furthermore, we develop a question generation approach while they regard the QA task as subgraph
search based on an integer linear programming (ILP) approach.
Our work also relates to \newcite{khashabi2018question}, which focuses on multi-choice question
answering based on the semantics of a document.
They use semantic role labeling and shallow parsing of a document to construct a semantic graph,
based on which an ILP based approach is developed to find the supporting subgraph.
The difference of our approach is that predicates from our document KB form are not limited to a predefined set,
so that they do not take into consideration the knowledge from external \mbox{KBs},
and also the difference in terms of methodology.
\newcite{miller2016key} answer questions based on \mbox{KBs} in the movie domain or
information extraction results from Wikipedia documents. Unlike this method, our approach focuses on entities from an external KB, our doc KB is obtained via open IE, and we combine the document KB with an open KB for question answering.

\section{Experiments}
We describe experiment settings and report figures and analysis in this section.

\subsection{Settings}
In our experiments, we tune model parameters on the development set and report results on the test set.
We design experiments from both ranking-based direction and question generation-based direction.
The evaluation metric is precision @1 \cite{bordes2014question}, which indicates whether the top ranked result is the correct answer.
We further report BLEU score \cite{papineni2002bleu} for the question generation approach.

We also adapt BiDAF~\cite{seo2016bidirectional}, a top-performing reading comprehension model on the SQuAD dataset~\cite{rajpurkar2016squad} as a strong baseline.
As BiDAF output a span from the input document as the answer to the given question,
we adapt it to KBMRC as a ranking model similarly as the approach used in previous research~\cite{semanticilp2018aaai}.
We use BiDAF to select an answer span from a corresponding document based on a given question
and select the candidate answer that has maximum overlap with the answer span as the final answer.

\subsection{Analysis: Question Answering Models}
Table~\ref{table:results-rank} shows the results of our two question answering models.
It is clear that KVMemNet achieves better P@1 scores on both dev and test sets than PCNet.
The reason is that candidate answers of PCNet come from the ``anchor'' point along 1-hop or 2-hop paths.
However, the correct answer might not be connected due to the quality of anchor detection.
On the dev set, we observe that only 69.6\% of correct answers can be covered by the set of candidate answers in PCNet, which  apparently limits the upper bound of the approach.
This is addressed in KVMemNet because all the arguments are candidate answers.
Both PCNet and KVMemNet outperform our implementation of \newcite{bordes2014question}, since the latter ignores word order.
We  incorporate each of the four KBs separately into PCNet and KVMemNet, and find that incorporating external KBs could bring improvements.

\begin{table}[!h]
	\centering
	\begin{tabular}{lcc}
		\hline
		{Method} &Dev&Test \\
		\hline
		\cite{bordes2014question}  &48.6&48.0\\
        BiDAF & 65.4& 62.3\\
		%		M-KBMRC (w/o) context &50.1 & 49.2\\
		\hline
		PCNet  & 50.7&49.4\\
		%		\hline
		\ \ + NELL     & 54.0&52.9\\
		\ \  + Reverb   & 54.1&52.6\\
		\ \  + Probase  & 54.8&\textbf{53.5}\\
		\ \  + Freebase & 54.7&53.1\\
		\hline
		KVMemNet  & 63.4& 63.6\\
		\ \  + NELL     & 64.8& 63.8\\
		\ \  + Reverb   & 64.4& \textbf{64.3}\\
		\ \  + Probase  & 64.0& 63.7\\
		\ \  + Freebase & 64.1& 63.7\\
		%\hline
        %BiDAF-sim1-F & 76.1& 74.2\\
        %BiDAF-sim1-B & 57.2& 55.8\\
        %BiDAF-sim2-F & 75.0& 72.9\\
        \hline
	\end{tabular}
	\caption{Performances (P@1) of different approaches on dev and test sets.}
	\label{table:results-rank}
\end{table}

From Figure~\ref{fig:kvmn_sample_2hops}, we can see that the KVMemNet model attends to the key ``(\textit{st. johns, is located})'' for the question ``\textit{Where is st johns mi located?}''.
Thus, the model has higher confidence in regarding value ``\textit{in clinton county}'' as the answer.

\subsection{Analysis: Generative Models}
Table~\ref{table:results-qg} shows the results of different question generation models.
Our approach is abbreviated as QGNet, which stands for the use of a paraphrasing model plus our question generation model.
We can see that QGNet performs better than Seq2Seq in terms of BLEU score because many important words of low frequency from the input are replicated to the target sequence.
However, the improvement is not significant for the QA task.
We also incorporate each of the four KBs into QGNet, and observe slight improvements on NELL and Reverb.
Despite the overall accuracy of QGNet being lower than PCNet and KVMemNet, combining outcomes with them could generates  1.5\% and 0.8\% absolute gains, respectively.

\begin{table}[t]
	\centering
	\begin{tabular}{llccc}
		\hline
		{Method}           &BLEU &P@1 \\
		\hline
		Seq2Seq            &11.2 & 41.3\\
		\hline
		QGNet             &16.3 & 41.5\\
		\ \   + NELL      &16.5 & 42.2\\
		\ \   + Reverb    &16.2 & 42.1\\
		\ \   + Probase   &16.5 & 41.8\\
		\ \   + Freebase  &16.1 & 41.3\\
		\hline
		QGNet + PCNet            &16.3 & 50.9\\
		QGNet + KVMemNet            &16.3 & 64.3\\
		\hline&
	\end{tabular}
	\caption{Performances of different question generation based systems on the test set. }
	\label{table:results-qg}
\end{table}

\begin{figure}[h]
	\centering
	\includegraphics[width=.47\textwidth]{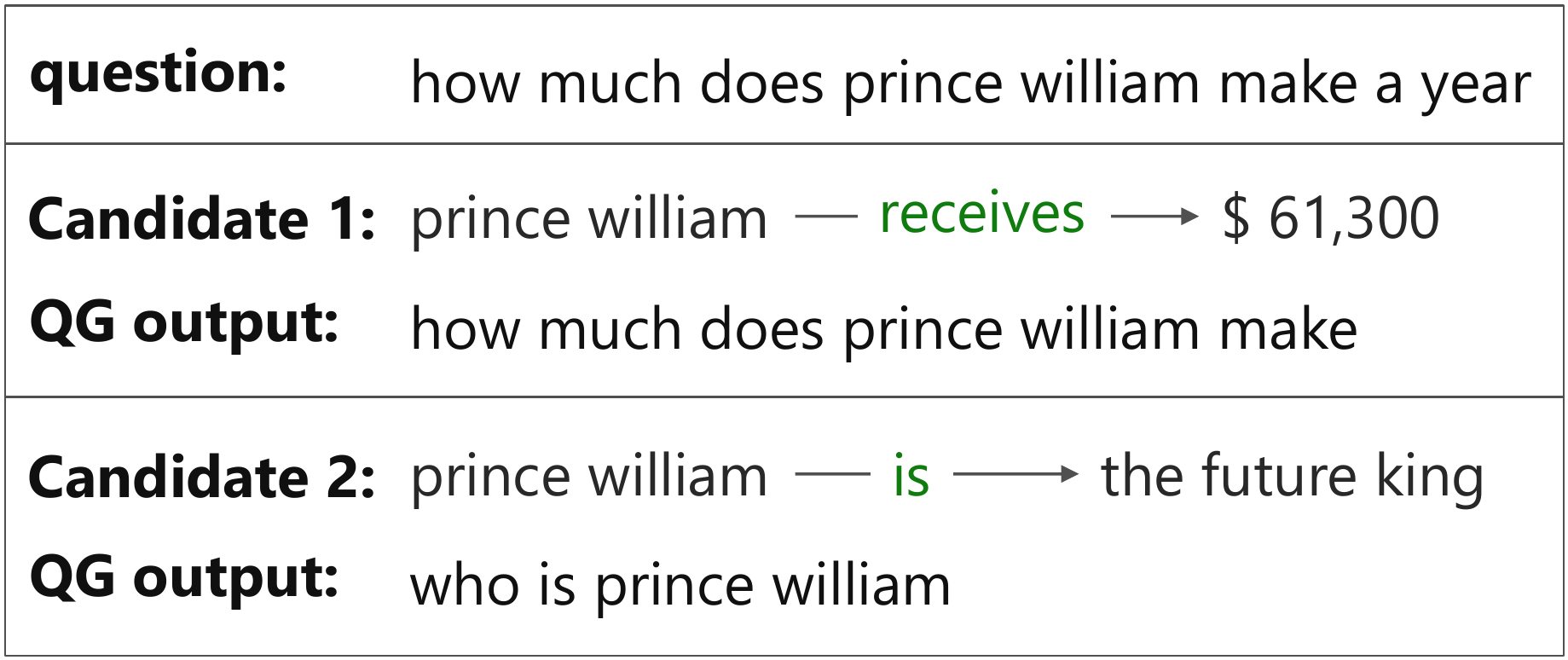}
	\caption{Example prediction of QGNet on the dev set, in which the first candidate is the correct answer while the second is incorrect. }
	\label{fig:qg-results}
\end{figure}

We show examples generated by our QG model in Figure \ref{fig:qg-results}, in which the paths of two candidate answers are regarded as the input to the QG model.
We can see that the original question is closer to the first generated result than the second one. Accordingly, the first candidate (\$ 61,300) would be assigned with a larger probability as the answer.

\section{Future Opportunities for Future Research}
We conduct error analysis from multiple perspectives to show the limitations and future opportunities of different components of our approach.

\paragraph{Question Answering Model.}
We study the limitations of KVMemNet by analyzing incorrectly predicted results.
We observe that the majority of errors are caused by mismatch between the question and the key of the correct answer.
One type of errors occurs when the subject is a pronoun.
For example,
the question ``\textit{What is the cost of caltech's solar-powered toilet}'' hardly matches to the
fact ``(\textit{it, will take, \$1,500 to \$2,000})'', in which the subject and predicate are the key and the object is the value.
Incorporating coreference resolution information might mitigate this issue.
Another type of error occurs when the key is a long sequence, in which the keyword  is a part of it but not well matched to the question.
Incorporating named entity recognition and leveraging it for question-key matching might be a potential direction for tackling this issue.

\paragraph{Question Generation Model.}
We analyze a randomly selected set of generated questions by QGNet, and categorize unsatisfied results into \mbox{two} groups.
The first group is generating duplicate words, which might be solved by incorporating a coverage mechanism.
The second group is replicating incomplete spans from the input. In this work, we do not design the model architecture to replicate an entire argument to the target sequence because a subject/objective might be too long.
However, word-level copying mechanism does not guarantee that a text span consisting of multiple words (such as named entity) will be successively copied.

\paragraph{External KB Retrieval.}
We observe randomly selected retrieved results from open KBs, and find that ElasticSearch performs pretty well in most cases that balances between accuracy and latency.
However, there still remain many cases in which named entities are partly matched, so that the retrieved results talk about totally different things. The problem can be mitigated by taking into consideration named entity information in both search queries and searchable values in ElasticSearch.
Despite the retrieved results being correct in terms of string match, some entities are ambiguous with our current strategy unable not distinguish between them.
For instance, the retrieved results for ``\textit{Louisiana}'' include ``(\textit{LOUISIANA, is a company in the economic sector of, services})'' and ``(\textit{LOUISIANA, is a state or province located in the geopolitical location, U.S.} )'', while only the second fact is the correct one.
Incorporating a disambiguation system is a potential solution to this problem.

\section{Conclusion}
In this paper, we focus on knowledge based machine reading comprehension.
We create a manually labeled dataset for the task, and develop a framework consisting of both question answering model and question generation model.
We further incorporate four open KBs as external knowledge into both question answering and generative approaches, and demonstrate that incorporating additional evidence from open KBs improves total accuracy.
We conduct extensive model analysis and error analysis to show the advantages and limitations of our approaches.
\bibliography{aaai2019}
\bibliographystyle{aaai}

\appendix

\end{document}